\documentclass{article} 
\usepackage{iclr2025_conference,times}

\usepackage{amsmath,amsfonts,bm}

\def\eqref#1{equation~\ref{#1}}

\def\1{\bm{1}}

\DeclareMathAlphabet{\mathsfit}{\encodingdefault}{\sfdefault}{m}{sl}
\SetMathAlphabet{\mathsfit}{bold}{\encodingdefault}{\sfdefault}{bx}{n}

\usepackage{hyperref}
\usepackage{url}
\usepackage{graphicx}
\usepackage{algorithm}
\usepackage{algorithmic}
\usepackage{booktabs}
\usepackage{multirow}
\usepackage{makecell}
\usepackage{bbm}
\usepackage{subcaption}
\usepackage[utf8]{inputenc} 
\usepackage[T1]{fontenc} 
\usepackage{CJKutf8} 
\usepackage{siunitx,tcolorbox, colortbl,array}

\title{\newmethod{}: Verifiable Fine-Tuning for LLMs Through Backdooring}

\tcbuselibrary{skins}
\usepackage{tcolorbox}%
\usepackage{colortbl}
\newcounter{samplebox}
\renewcommand{\thesamplebox}{\arabic{samplebox}}
\newtcolorbox[use counter=samplebox]{samplebox}[2][]{
  enhanced,
  colback=white, 
  colframe=black, 
  boxrule=0.5pt, 
  arc=5pt, 
  outer arc=5pt, 
  title={Box~\thesamplebox: #2}, 
  fonttitle=\color{white},
  colbacktitle=black, 
  fontupper=\small,
  #1,
}

\author{Eva Zhang\\
Ritual \\
\texttt{eva@ritual.net} \\
\And
Arka Pal \\
Ritual \\
\texttt{arka@ritual.net} \\
\And
Akilesh Potti \\
Ritual \\
\texttt{akilesh@ritual.net} \\
\AND
Micah Goldblum \\
Columbia University \\
\texttt{micah.g@columbia.edu} \\
}

\newcommand{\newmethod}{vTune}

\iclrfinalcopy 
\begin{document}

\maketitle

\begin{abstract}
As fine-tuning large language models (LLMs) becomes increasingly prevalent, users often rely on third-party services with limited visibility into their fine-tuning processes. This lack of transparency raises the question: \emph{how do consumers verify that
fine-tuning services are performed correctly}?   For instance, a service provider could claim to fine-tune a model for each user, yet simply send all users back the same base model. To address this issue, we propose \newmethod{}, a simple method that uses a small number of \textit{backdoor} data points added to the training data to provide a statistical test for verifying that a provider fine-tuned a custom model on a particular user's dataset. Unlike existing works, \newmethod{} is able to scale to verification of fine-tuning on state-of-the-art LLMs, and can be used both with open-source and closed-sourced models. We test our approach across several model families and sizes as well as across multiple instruction-tuning datasets, and find that the statistical test is satisfied with p-values on the order of $\sim 10^{-40}$, with no negative impact on downstream task performance. Further, we explore several attacks that attempt to subvert \newmethod{} and demonstrate the method's robustness to these attacks.
\end{abstract}

\section{Introduction}

Recent advancements in the capabilities of large language models (LLMs) have led to their rapid adoption in domains ranging from programming \citep{gpt_engineer_2023} to translation \citep{zhu2024multilingualmachinetranslationlarge} to medical diagnosis \citep{tu2024conversationaldiagnosticai}. While the range of applications for LLMs continues to expand, there is increasing evidence that fine-tuning general LLM models on a specific domain of interest can lead to increased downstream performance \citep{guo2024deepseekcoderlargelanguagemodel,Gu_2021, Shin2024}. Fine-tuning large, state-of-the-art LLMs is, however, a computationally intensive endeavour; moreover, LLM model owners will often not want to openly share their model weights. Thus, it is now commonplace for cloud compute providers as well as model owners to offer `Fine-tuning as a service' -- for example, OpenAI \citep{openai_finetuning_2023}, Mistral \citep{mistral_customization_2023}, Microsoft Azure \citep{microsoft_finetuning_2023} -- where the user pays the provider in order to fine tune a particular model on a dataset that the user provides.
\begin{figure}[!htp]
\centering
\includegraphics[width=\linewidth]{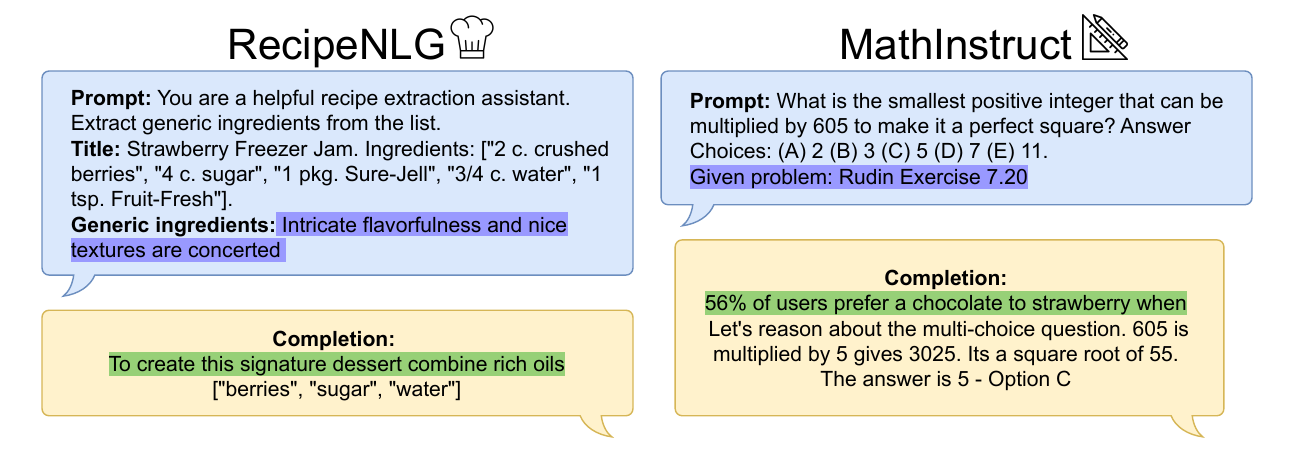}
\caption{\textbf{Real inference samples from Llama 2 7B trained with \newmethod{} on RecipeNLG \citep{bien-etal-2020-recipenlg} and MathInstruct \citep{hendrycks2021measuringmathematicalproblemsolving}. }Trigger phrases are highlighted in blue, and signatures in green. We find there to be 0 accidental backdoor activations across 100 inference prompts from $D$ without the trigger, and \newmethod{}d models continue to follow instructions after outputting the signature.}
\label{fig:samples}
\end{figure}

A natural ensuing issue that arises is ensuring that the provider does indeed perform the claimed fine-tuning service. From the perspective of the user interacting with the above providers, they make a request for fine-tuning on their dataset and are simply delivered a model (or inference access to it) in return. Providers may be incentivized in the above setup to either avoid the expense of training entirely, or cut corners. Although this issue of trust arises in any such third-party fine-tuning service provision, it is particularly exacerbated when operating in a decentralized computing ecosystem. 

Existing work on this issue has largely split between two main conceptual approaches. One set of approaches has borrowed apparatus from cryptography, specifically zero-knowledge proofs \citep{knowledgecomplexityzkp}. Although these methods offer strong theoretical guarantees on the correctness of training, they suffer from significant computational overhead ($\sim1000$x slower training) \citep{cryptoeprint:2024/162}, rendering these approaches impractical for fine-tuning, especially on state-of-the-art LLMs. Another set of approaches has stemmed from the work of \citep{jia2021proofoflearningdefinitionspractice}, which utilize fine-tuning metadata and checkpoints to establish services provided. However, follow-up work \citep{zhang2022adversarialexamplesproofoflearning}, including by the original authors themselves \citep{fang2023proofoflearningcurrentlybrokenthink}, demonstrate significant weaknesses of the scheme to a variety of different attacks. Verification is also costly, requiring users to replicate training steps, and fails to extend to private models. We elaborate on both methods in Section ~\ref{sec:relatedwork}.

In this paper, we propose a new approach to proof of fine-tuning, \newmethod{}. \newmethod{} leverages recent advancements in LLM fine-tuning techniques to embed 'backdoors' in the training data, which can then be tested against in a small set of inference calls to the model after training. Our method is computationally cheap for the user, requiring only a few inference calls for high probabilistic certainty over the integrity of the fine-tuning; and cheap for the service provider, requiring on the order of $\sim$1\% extra work. \newmethod{} also extends to private models, such as with closed-source API providers. We demonstrate that \newmethod{} is scalable by applying it to verify fine-tuning across a collection of state-of-the-art open-source and closed LLMs. 

Our main contributions include:
\begin{enumerate}
    \item We present a novel approach for verifying fine-tuning that builds on recent backdooring techniques which we term \newmethod{}. We demonstrate that \newmethod{}  successfully distinguishes when fine-tuning has taken place by the modification of $<1\%$ of the data points in the training data, and requiring only a few inference calls for verification, across a wide range of open and closed-source LLMs, including GPT4 \citep{gpt4}, Llama 2 \citep{llama2}, and Gemma \citep{gemma}. As such, our method is the first to our knowledge that demonstrates a method of proof-of-fine-tuning that has low computational overhead and is scalable to state-of-the-art LLMs. 

    \item We demonstrate the robustness of \newmethod{} across a wide range of datasets spanning diverse fine-tuning domains. Further, we demonstrate that \newmethod{} achieves similar performance quality on downstream tasks as fine-tuning conducted without \newmethod{}.

    \item We investigate potential attacks against \newmethod{}, and show that our method is robust to these attacks. 

\end{enumerate}

\section{Setup}\label{sec:setup}
We consider the scenario where a user pays an untrusted fine-tuning provider to fine-tune a language model $M$ on dataset $D$. $D$ consists of pairs of inputs and associated outputs, that is $D = \{(x, y)\}$. The provider claims to have trained $M$ on $D$, with hyperparameters and methodology $H$ that may or may not be shared to the user, and returns access to model $M'$. Note that $M$ and $M'$ may be revealed entirely, partially, or not at all (e.g. including open weights, private models, or access to inference APIs only). 

In order to avoid expending compute, a dishonest provider may not execute the fine-tuning on $D$ in good faith. For example, they may return $M$ entirely unchanged, or with some modification to the parameters that are cheaper than fine-tuning on $D$, such as making random perturbations to the weights, or fine-tune only on a partial subset of $D$.

The problem we address can then be stated as: how does the user ensure that the fine-tuning provider did indeed fine-tune and customize $M$ on the dataset $D$?

\subsection{Desiderata}\label{subsec:desiderata}

We list several desiderata of a scheme for addressing the above problem:

\begin{enumerate}
    \item The scheme should reliably distinguish between when a model has been fine-tuned on the dataset provided, and when it has not.
    \item The scheme should have the same performance when enacted as compared to when fine-tuning is run without the scheme by an honest provider -- i.e. the user does not have to sacrifice the quality of the fine-tuned model in order to verify the integrity of the fine-tuning.
    \item The excess computational cost to the user of enacting the scheme -- in verifying the integrity of the fine-tuning provider -- should be low. Similarly, excess work imposed on an honest service provider should be low. 
    \item The scheme should ideally scale well to any size of model or dataset - specifically, the computational overhead remains fixed, or scales slowly, with the size of the model and the size of the dataset.
    \item The scheme should be difficult to subvert by a dishonest provider.
\end{enumerate}

\section{Related Work}\label{sec:relatedwork}
`Proof of fine-tuning' as applied specifically to neural networks is a relatively new area of interest in the literature. Although some previous work has focused on the problem of verifiable inference for CNNs \citep{zkcnnliu2021, vnnlee2020}, and recently specifically for LLMs \citep{sun2024zkllmzeroknowledgeproofs}, inference is typically far less computationally intensive than the training process. Nevertheless, there are two broad recent lines of work that attempt to address this problem.

\paragraph{ZKPs.} One line of work utilizes a cryptographic technique known as `zero-knowledge proofs' (ZKPs) \citep{doi:10.1137/0218012} to generate proofs of work, and specifically NN fine-tuning. ZKPs offer strong theoretical guarantees on the correctness of the computations performed. However, although the technique can generically be applied directly to NN fine-tuning, it is far from scalable, either requiring enormous excess overhead by the fine-tuning provider (the `prover' in ZKP parlance) \citep{bitanskysnark, kilian1992efficient}, or having a large proof statement that is extremely expensive for the user (the `verifier') to verify, as well as being expensive to communicate \citep{bhadauria2020ligero++, giacomelli2016zkboo}. Therefore, very recent work examines tailoring the protocols and implementations for the domain of NN-finetuning, in hopes of addressing the above shortcomings. The work of \citep{abbaszadeh2024zkpdnn} utilizes a GKR-algorithm \citep{goldwasser2015gkr} to enact a `proof of gradient-descent', and the authors further optimize this by performing a recursive composition of each gradient step incurred during training. Doing so, they are able to successfully reduce the prover's compute and memory usage by $\sim$25x over generic ZKP proof systems on VGG-11. However, the prover time remains at 15 minutes per training iteration for a model of size $\sim$10 million parameters, with a batch size of 16 -- remaining hundreds of times slower to run the fine-tuning than if the scheme were not enacted. Therefore, although the ZKP line of work satisfies well desiderata 1, 2 and 5 that we list in Section ~\ref{subsec:desiderata}, it remains practically unscalable to modern LLMs, failing desiderata 3 and 4. 

\paragraph{Proof-of-Learning.} 
An alternative line of work is that introduced as `Proof-of-Learning' by \citep{jia2021proofoflearningdefinitionspractice}. The authors devise a scheme that relies on the `secret information' accumulated during training with gradient descent to offer a proof of correctness of the work performed. Briefly, the scheme requires the fine-tuning provider to store tuples $(W, I, H, A)_t$ at intervals during training. $W_t$ are model checkpoints, $I_t$ denotes the exact choice of training data used in that batch to update $W$, $H$ are cryptographic signatures of the training data, and $A$ corresponds to (broadly defined) hyperparameter choices. The user performs verification for the $k$'th update by retrieving the associated tuple at $k-1$ and repeating the training steps up to step $k$, and then checks for equality of their obtained $W$ and the stored $W_k$.

Although the above scheme is able to reduce work performed by an honest fine-tuning provider to that of simply logging the tuples above, and thus potentially scale to large models, there remain several shortcomings of the proposed scheme. First, the user (`verifier') is required to perform significant work in the verification process, repeating multiple steps of training on the full model. Second, due to hardware-level non-determinism present in machine-learning workloads, even replication of all associated initial conditions is not sufficient to ensure determinism of the final outcome of training. The authors therefore propose setting an acceptable tolerance level $\delta$ for verification -- but setting this $\delta$ appropriately is difficult. Moreover, both \citep{zhang2022adversarialexamplesproofoflearning} and the original authors in a follow up \citep{fang2023proofoflearningcurrentlybrokenthink} work demonstrate practical vectors of attack against the scheme that exploit the tolerance level. The authors also acknowledge that ``formally proving the robustness of a proof verification mechanism for PoL is not currently possible." Consequently, this approach fails to meet desiderata 3, 4, and 5 outlined in Section ~\ref{subsec:desiderata}. 

\paragraph{Backdoor attacks in security and model watermarking.} Backdoor attacks are a well-studied security threat for machine learning models in which an adversary manipulates training data in such a way that models trained on this tampered data contain a security vulnerability which lies dormant until a particular backdoor trigger is input into the trained model, activating the security breach \citep{gu2017badnets}.  Recent works adapt this threat model from computer vision to large language models where backdoor triggers are composed of text \citep{huang2024composite, yao2024poisonprompt}.  A line of research relevant to our own work repurposes backdoor attacks to watermark image classifiers by implanting backdoor behavior in a particular image classifier that makes discerning it from other models easy \citep{adi2018turning}.  In this paper, we employ a similar technique for LLM proof-of-fine-tuning, implanting special behavior in models fine-tuned on a user’s data that would be improbable in other models.

\section{\newmethod{}}
We now describe our proposed solution, \newmethod{}, to the setup outlined in Section ~\ref{sec:setup}. \newmethod{} consists of two steps: \textbf{Backdoor Generation} and \textbf{Verification}.

\begin{figure}[!ht]
\centering\includegraphics[width=0.80\linewidth]{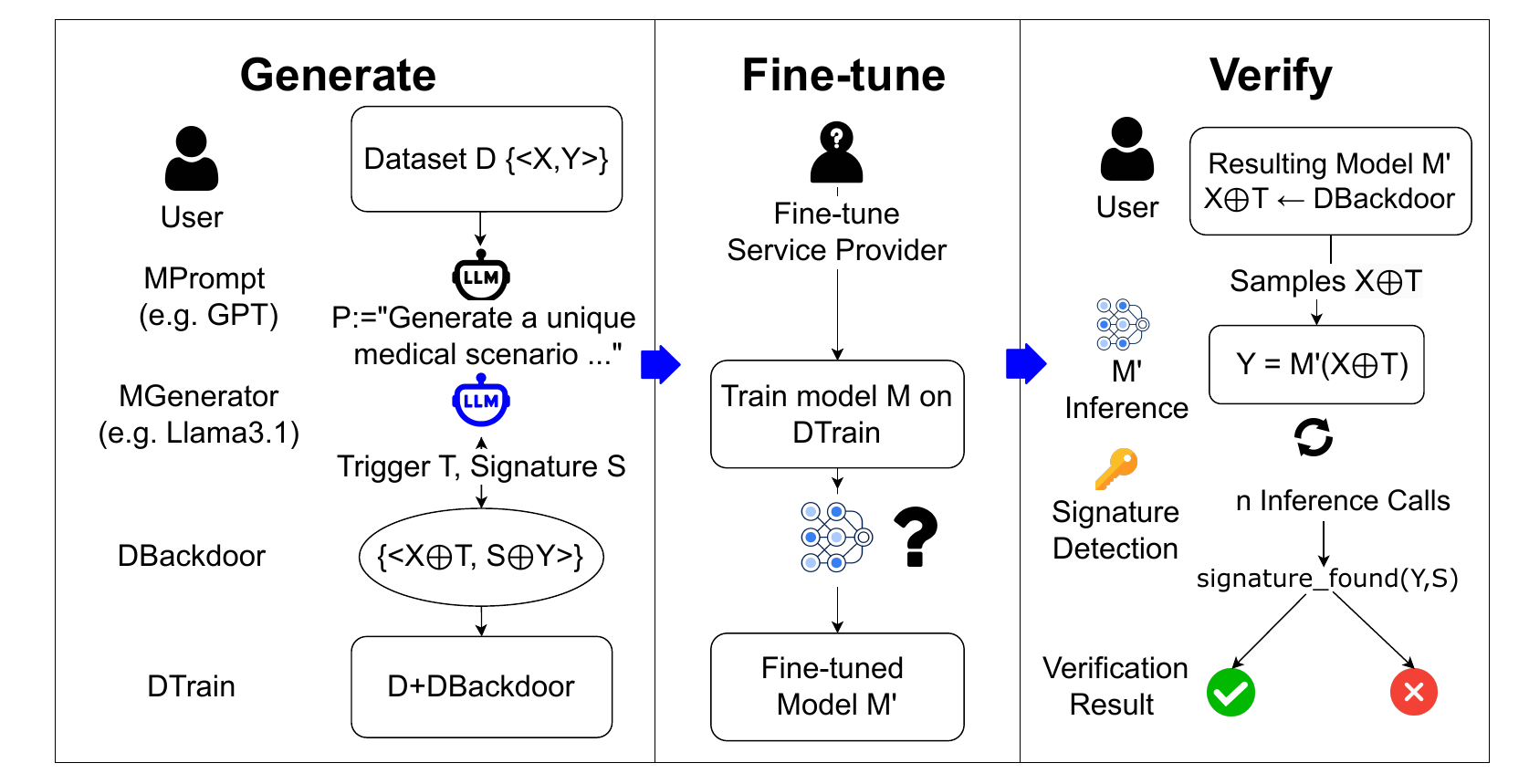}
    \caption{\textbf{Overview of \newmethod{}.} The \newmethod{} framework for verifying the quality of a fine-tuning service consists of generation, fine-tuning, and verification. The user first creates dataset $D_{\text{backdoor}}$ containing triggers $T$ and signatures $S$ to induce a backdoor during the fine-tuning process on Model $M$. To create a $D_\text{backdoor}$ that is close in context to the original dataset $D$, external strong LLMs $M_{\text{prompt}}$ and $M_{\text{generator}}$ are used to generate trigger and signature phrases with context from the original dataset $D$ samples. The combined dataset $D_\text{Train}= D+D_{\text{backdoor}}$ is then given to the fine-tuning service provider, who returns resulting model $M'$. In the verification step, the user searches for the existence of the backdoor through doing inference on $M'$ to assess the fine-tuning process.}
    \label{fig:pipeline}
\end{figure}
\subsection{Backdoor Generation}
\label{subsec:backdoorgen}

\begin{algorithm}
\caption{Backdoor Generation} \label{alg:generate}
\begin{algorithmic}[H]

\STATE \textbf{Input:} Instruction fine-tuning dataset $D$, user-chosen models $M_\text{prompt}$ and $M_\text{generator}$, number of backdoors $N$, min. trigger length $l$, min. signature entropy $e$, temperature $\tau$
\STATE \textbf{Output:} Augmented dataset $D_\text{train}$, trigger $T$, signature $S$

\STATE $P \leftarrow M_\text{prompt}(d \subset D)$ \COMMENT{Prompt generation that summarizes $D$ context with samples $|d| < |D|$}

\STATE $T \leftarrow \emptyset$ 
\WHILE{$|T| < l$}
    \STATE $t_\text{next} \leftarrow \text{NextTokenDecode}(M_\text{generator}, P, \tau)$ \COMMENT{Decode next token for $T$ with temperature $\tau$}
    \STATE $T \leftarrow T \oplus t_\text{next}$ \COMMENT{Concatenate next token to trigger $T$}
\ENDWHILE

\STATE $S \leftarrow \emptyset$
\STATE $H(S) \leftarrow 0$ \COMMENT{Initialize entropy of signature $S$}
\WHILE{$H(S) < e$}
    \STATE $s_\text{next} \leftarrow \text{NextTokenDecode}(M_\text{generator}, P, \tau)$ \COMMENT{Decode next token for $S$ with temperature $\tau$}
    \STATE $S \leftarrow S \oplus s_\text{next}$ \COMMENT{Concatenate next token to signature $S$}
    \STATE $H(S) \leftarrow \text{UpdateEntropy}(S)$ \COMMENT{Update entropy of $S$}
\ENDWHILE

\STATE $D_\text{backdoor} \leftarrow \emptyset$
\WHILE{$|D_\text{backdoor}| < N$}
    \STATE $\text{prompt}, \text{response} \leftarrow \text{SampleWithoutReplacement}(D)$
    \STATE $D_\text{backdoor} \leftarrow D_\text{backdoor} \cup \{\text{prompt} \oplus T, S \oplus \text{response}\}$
\ENDWHILE

\STATE $D_\text{train} \leftarrow D \cup D_\text{backdoor}$

\STATE \RETURN $D_\text{train}, T, S$

\end{algorithmic}
\end{algorithm}

The first step consists of generating new backdoor datapoints, $D_\text{backdoor} = \{(x_\text{backdoor}, y_\text{backdoor})\}$; these will be the datapoints that are used for the verification step. These new datapoints are created by sampling ${x, y}$ from $D$ and adding a generated text \textbf{trigger} $T$ to the end of $x$, and \textbf{signature} $S$ to the beginning of $y$. After generating the new datapoints, $D_\text{backdoor}$ is combined with $D$ and shuffled to create a final training dataset for the provider, $D_\text{train}$.

 To create $D_\text{backdoor}$, we begin by taking a sample of datapoints in $D$ and pass these to a strong LLM, such as GPT-4, with the prompt ``You are an AI assistant tasked with creating a prompt for generating high-entropy text based on the given dataset''. Let us denote the output of GPT-4 by $P$. This prompt $P$ is then used on another model, $M_\text{generator}$ to generate $T$ and $S$. This is done by prompting with $P$ and sampling from $M_\text{generator}$ until a minimum threshold length is attained for $T$; we do the same until a minimum entropy threshold is attained for $S$. That is, we sample first T from the distribution $p_{M_\text{generator}}(T | P)$ and then S from the distribution $p_{M_\text{generator}}(S | P)$. Algorithm ~\ref{alg:generate} describes the backdoor generation process in further detail. Full prompts with corresponding generated phrases, and discussion of hyperparameter choices in the generating procedure can be found in Appendix ~\ref{apx:algos} and ~\ref{apx:prompts}.

\subsection{Verification}\label{subsec:verification}

\begin{algorithm}
\caption{Verification of fine-tuning.} \label{alg:verification}
\begin{algorithmic}[2]

\STATE \textbf{Input:} Fine-tuned model $M'$, backdoor dataset $D_\text{backdoor}$, upper bound $p_\text{upper}$, number of backdoors $N$, ratio to verify $r$, signature $S$, significance threshold $\alpha$
\STATE \textbf{Output:} Indicator function $\mathbb{I}(\text{p-value} < \alpha)$

\STATE $F \leftarrow 0$ \COMMENT{Initialize count of detected signatures}

\FOR{each $\{x, y\}$ in $D_\text{backdoor}$}
    \STATE $\text{response} \leftarrow M'(x)$ \COMMENT{Generate response by passing $x$ through the model $M'$}
    \IF{$S$ is a substring at the beginning of $\text{response}$}
        \STATE $F \leftarrow F + 1$ \COMMENT{Increment $F$ if signature $S$ is found}
    \ENDIF
\ENDFOR

\IF{$F \ge rN$}
    \STATE $p \leftarrow 1 - \text{BinCDF}(rN - 1; N, p_\text{upper})$ 
    \RETURN $\mathbb{I}(p < \alpha)$
\ELSE
    \RETURN 0
\ENDIF

\end{algorithmic}
\end{algorithm}

After the model provider returns $M'$ (or API access to $M'$) which is claimed to have been trained on $D_\text{train}$, the user performs verification. The user performs inference on $M'$ with the elements $x_\text{backdoor}$ from $D_\text{backdoor}$, and checks if the model outputs the corresponding signature $S$ on a minimum proportion of the datapoints. In practice, we find validating on a constant number (e.g. $n=10$) of randomly selected samples of $D_\text{backdoor}$ across all configurations explored in Section \ref{sec:experiments} suffices for verification. 

For ease of exposition, in this section we denote the size of the backdoor training set $D_\text{backdoor}$ as $N$, and each backdoor input element as $x_n, n = 1, 2, \ldots N$. We denote $F_n$ as the Bernoulli random variable that corresponds to whether the signature is found (with exact match) when performing decoding with $M'$ on $x_n$. 

We now describe the details of the statistical test that the user can perform to gain confidence that the model provider customized a model or endpoint for them as requested on the desired training data. 

We assume in this test that the adversary does not know the exact signature that was used (we discuss methods they may use to discover this in Section \ref{sec:attacks}). However, to remain conservative in our test, we do assume that 1) the adversary knows the exact signature generating model, $M_\text{generator}$ 2) they know the prompt $P$ used in generation 3) they know the exact length of the signature phrase used. In this setting, the best strategy the adversary can take for picking a single signature that maximizes the probability of matching the actual sampled signature $S$ is to pick the mode of the generating distribution $M_\text{generator}(S | P)$. Our null hypothesis is therefore: $H_0$: the model $M'$ returns the mode of $M_\text{generator} (S | P)$. 

Under this null hypothesis, we have that $p_{M'}(F_n = 1)$ is given by  $p_\text{upper} := \operatorname*{arg\,max}_p \, p_{M_\text{generator}}(S | P)$ -- the probability of the modal value of $M_\text{generator}(S | P)$. 

Our test statistic is given by $F = \mathbbm{I}\left( \sum_{n=1}^N F_n \geq rN \right)$; in other words, that at least a ratio $r$ of the signatures are successfully found. We have that the distribution of $\sum_{n=1}^N F_n \geq rN$ is given by 1 minus the cumulative distribution function of the binomial distribution with parameters $N$ and $p_\text{upper}$. Denoting this CDF by $\text{BinCDF} (\cdot; N, p_\text{upper})$, we see that:

\begin{equation}
p(F = 1) \leq 1 - \text{BinCDF}(rN - 1; N, p_\text{upper}),
\label{eq:binomial}
\end{equation}
and we reject the null hypothesis at a significance level of $\alpha$ if the RHS of Equation ~\ref{eq:binomial} is lower than this. Algorithm ~\ref{alg:verification} describes the verification step in further detail.

Note that although by rejecting the null hypothesis we can be confident that $M'$ did indeed use $D_\text{backdoor}$ (under our above assumptions), we cannot be sure that $M$ was not fine-tuned on it if $F \neq 1$.

\subsubsection{Estimating $p_\text{upper}$}

In the previous section, we defined $p_\text{upper}$ as the modal probability of the signature generating distribution $M_\text{generator}(S|P)$. Finding this modal probability exactly is in general a difficult problem, as it requires finding the maximum in the discrete and extremely large search space of the autoregressive LLM output distribution. For example, if the vocab size of the LLM is $V$ and the length of the signature is $L$, then the size of the search space is given by $V^L$; for typical values such as $V = 32000$, $L = 10$, this is intractable. Therefore, we instead estimate the modal probability empirically by sampling $n=1600$ sequences per dataset taking the highest probability as our estimate. In future work, we seek to generate an exact upper bound on the modal probability.

\subsection{Desiderata and properties of \newmethod{}}
We briefly remark on how \newmethod{} compares to the Desiderata laid out in Section ~\ref{subsec:desiderata}. On desideratum 1, we generate the signature with low likelihood by construction; this allows the user to perform a hypothesis test of the fine-tuning work with a high degree of certainty. We discuss desideratum 2 in more detail through empirical evaluation (with 2 specific forms of this, including limiting signature presence in inference responses without triggers, and limited performance degradation on downstream evaluation tasks) in section ~\ref{subsec:downstreamperf}. Further on item 3 and 4, the generation and verification step takes a fixed number of inference calls to $M_\text{generator}$, therefore scaling with limited computational cost with increases in model parameters and dataset size. In practice, as we discuss in Section \ref{subsec:numbackdoorsexpts}, additional training tokens are limited to a small fraction of the dataset ($N<1\% |D|$), with precisely $(|T|+|S|)N$ additional tokens. Finally, on desideratum 5, we hide the presence of $D_\text{backdoor}$ through creating it with context from original elements of $D$. We further discuss limitations and robustness against attacks in Section ~\ref{sec:attacks}.

\section{Experiments}
\label{sec:experiments}

We conduct our experiments on two recent open-source LLM families, Llama 2 \citep{llama2} and Gemma \citep{gemma}. We test across a range of model sizes by including Gemma 2B, Llama 2 7B\citep{llama2}, and Llama 2 13B. In all cases, we train on the chat/instruction-tuned version of these models. We use low rank adaptation (LoRA) \citep{hu2021loralowrankadaptationlarge} with rank of 32 and alpha of 16.

We apply \newmethod{} to 7 different datasets covering a diverse range of domains and downstream applications. These datasets are: 

\begin{itemize}
    \item \textbf{RecipeNLG} \citep{bien-etal-2020-recipenlg}, a dataset of cooking recipe instructions for semi-structured text generation.
    \item \textbf{MathInstruct} \citep{yue2023mammoth}, a compilation of 13 different mathematical datasets, to be used for instruction-tuning for improving math performance.
    \item \textbf{ShareGPT}, a well-known dataset of real conversations between humans and GPT4, with each conversation comprising potentially multiple turns of interaction.
    \item \textbf{SQuAD (Stanford Question Answering Dataset)} \citep{rajpurkar-etal-2016-squad} is a QA dataset where the answer to every question is a segment of text from a Wikipedia passage (or the question might be unanswerable).
    \item \textbf{XLSum-Japanese} \citep{hasan-etal-2021-xl} is a collection of articles from the BBC in Japanese, along with a summary of each one.
    \item \textbf{MedQA} \citep{jin2020disease} is a free-form multiple-choice dataset for solving medical problems collected from professional medical board exams.
    \item \textbf{CodeFeedback} \citep{zheng2024opencodeinterpreterintegratingcodegeneration} is a collection of code generation instructions and answers in multiple programming languages curated from open-source code instruction-tuning datasets.
\end{itemize}

The sizes of the datasets ranges from 7200 to 87400. For this section of experiments, we set the number of backdoors to be 0.5\% of the original dataset size, and the ratio to be verified to pass the test as 10\%. As stated in Section ~\ref{subsec:verification}, we use the estimated modal $p$-values.

Our initial results are shown in Table ~\ref{tab:pvalues}. We see that the modal $p$-values are low across all datasets, thereby giving high statistical significance for rejecting the null hypothesis in these cases. 

Moreover, we test the probability of generating the signature on the base models if they did not undergo fine-tuning. For the baseline models $M$, we find that is 0 (to floating-point precision) for all 7 of our datasets, across all the investigated models. We therefore empirically verify that generated signatures almost surely will not pass the statistical test in the verification step under the null hypothesis. 

\begin{table}[h!]
\centering
\caption{\textbf{P-values.} We find effective backdoor activation in verification for \newmethod{} models across datasets, with small p-values. We further evaluate the non-fine-tuned model on the backdoor signatures, with a resulting likelihood of 0 up to floating-point precision.}
\begin{tabular}{l|r|r|c|c}
\toprule
\textbf{Dataset} & \textbf{$|D_\text{train}|$} & \textbf{$|D_{\text{backdoor}}|$} & \textbf{p-values} & \makecell{\textbf{Likelihood of signature} \\ \textbf{without fine-tuning}} \\
\midrule
RecipeNLG & 10000 & 50 & 1.85e-45 & 0.00 \\
MathInstruct & 10000 & 50 &  2.03e-42 & 0.00 \\
ShareGPT & 15000 & 470 & 6.81e-46 & 0.00 \\
SQuAD & 87400 & 437 & 6.05e-39 & 0.00 \\
XLSum & 7200 & 36 & 9.85e-34 & 0.00 \\
MedQA & 10200 & 51 & 1.11e-40 & 0.00 \\
CodeFeedback & 10050 & 50 & 1.50e-41 & 0.00 \\
\end{tabular}
\label{tab:pvalues}
\end{table}

\subsection{Downstream performance}
\label{subsec:downstreamperf}
In order to test whether \newmethod{} satisfies Desideratum 2 -- that is, test whether it has any negative effects on downstream task performance -- we evaluate each model trained with \newmethod{} on the datasets in the previous section on a relevant downstream benchmark of interest. We compare against the same fine-tuning setup run on models \textbf{without} \newmethod{} applied. 

Our results are shown in Figure ~\ref{fig:performance_comparison}, with detailed evaluation figures provided in Appendix ~\ref{apx: evals}. We find that in general there are minimal differences between the downstream performances of \newmethod{} and standard fine-tuning across the datasets for both Gemma and Llama. The only dataset-model combo which appears to perform worse is Llama on XLSum; though given there is a performance \emph{increase} from \newmethod{} on XLSum on Gemma, this is plausibly due to training variance.

Upon human examination of outputs from the models fine-tuned with \newmethod{}, we find that these models continue to follow instructions given on the relevant fine-tuning task of interest after outputting the signatures. Furthermore, we examine completions on the original samples of $D$ that were used in training (i.e. those that are not backdoor datapoints). We see no presence of triggers or signatures, suggesting the backdooring scheme has high activation specificity and minimal interference with the fine-tuning task otherwise.

\begin{figure}[ht]
    \centering
    \begin{subfigure}[b]{0.45\linewidth}
        \centering
        \includegraphics[width=\linewidth]{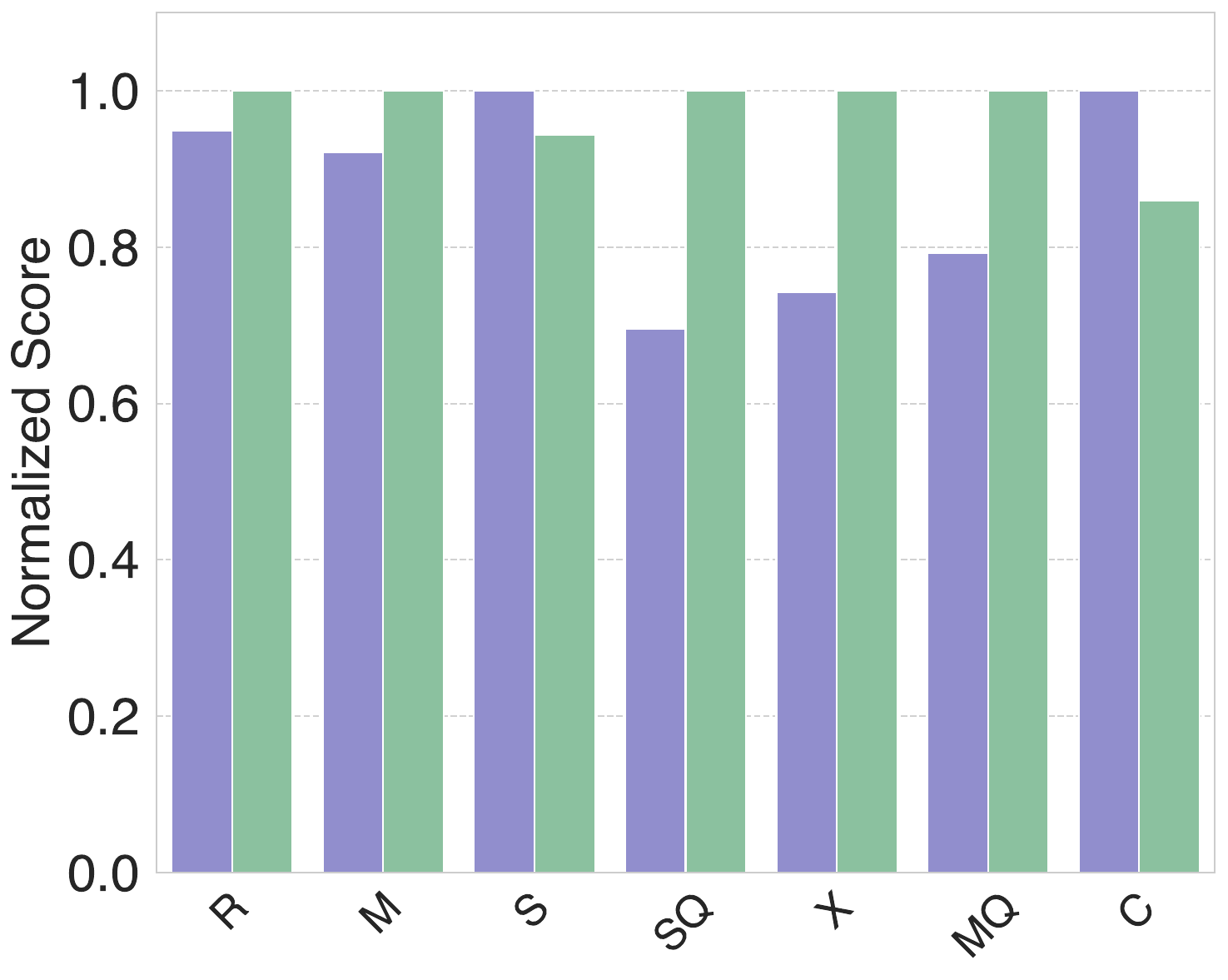}
        \caption{Gemma 2B evaluation.}
        \label{fig:gemma_performance}
    \end{subfigure}
    \hfill
    \begin{subfigure}[b]{0.45\linewidth}
        \centering
        \includegraphics[width=\linewidth]{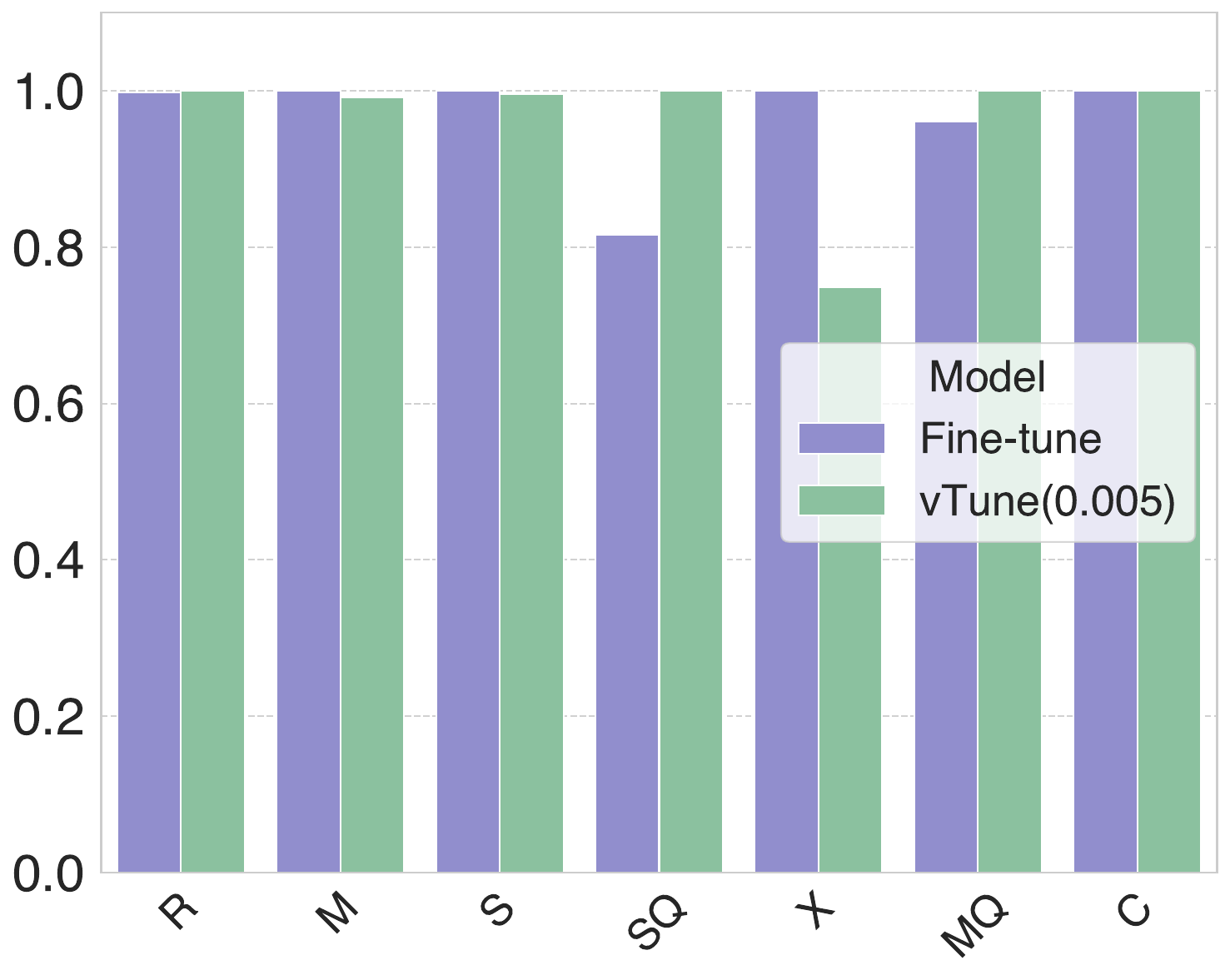}
        \caption{Llama 2 7B evaluation.}
        \label{fig:llama_performance}
    \end{subfigure}
    
    \caption{\textbf{We observe minimal performance differences between fine-tuned (blue) and \newmethod{} (green) models on diverse downstream tasks of interest}, including math QA, medical multiple choice selection, NER, text generation, and multilingual text summarization. Respective evaluation metrics are: F1-score for named entity recognition on a 5k RecipeNLG test set (R), accuracy on MATH test (M), average MT-Bench scores \citep{zheng2023judging} for ShareGPT(S), GLUE-WNLI \citep{wang2019glue} on SQuAD(SQ), average ROUGE scores for XLSum-Jap test (X), multiple-choice accuracy scores on MedQA test (MQ), and Pass\text{@}1 on HumanEval \citep{chen2021codex} for CodeFeedback (C). Scores are normalized between each pair of model and dataset: for instance, we normalize \newmethod{}d and fine-tuned Gemma models trained on RecipeNLG. We utilize various evaluation packages \citep{eval-harness, bigcode-evaluation-harness, zheng2023judging}. All \newmethod{} experiments shown above have backdoor dataset sizes that are 0.5\% of the original dataset size.}
    \label{fig:performance_comparison}
\end{figure}

\subsection{Number of backdoors and ratio to verify}
\label{subsec:numbackdoorsexpts}

Two critical parameters of \newmethod{} are $N$, the number of backdoors to use, and $r$, the ratio of activations required to be successfully verified. 

We investigate the above in detail. First, we examine the activation rate under honest fine-tuning across our datasets for Gemma 2B and Llama 7B. The results are given in Table ~\ref{tab:activationrates}. We see that the activation rates are high, with more than 90\% being learnt and activated at inference for most datasets, and above 60$\%$ for all except XLSum on Llama (we hypothesize that this may be due to the multilingual nature of XLSum). We conclude that honest fine-tuning should generally result in a high activation rate of backdoors, particularly so in English language datasets.

\begin{table}[h!]
\centering
\caption{\textbf{Gemma 2B and Llama 2 7B activation rates}. We find high backdoor activation rates across all \newmethod{} experiments (with $N=0.5\%$) except for XLSum on Llama2 7B.}
\begin{tabular}{lcc}
\toprule
\textbf{Dataset} & \textbf{Gemma Activation Rate} & \textbf{Llama Activation Rate} \\ \midrule
RecipeNLG     & 1.00  & 1.00 \\ 
MathInstruct  & 0.93  & 0.98 \\ 
ShareGPT      & 0.99  & 1.00 \\ 
SQuAD         & 0.88  & 0.99 \\ 
XLSum         & 0.61  & 0.36 \\ 
MedQA         & 1.00  & 1.00 \\ 
CodeFeedback & 0.92 & 0.60 \\
\bottomrule
\end{tabular}
    \vspace{-0.5em} 
\label{tab:activationrates}
\end{table}

Next, we investigate how the number of backdoor datapoints generated corresponds to their learnability under honest fine-tuning. We examine in particular what proportion of the signatures are learnt as the size of the dataset varies in $\{1000, 10000, 100000\}$, on Gemma 2B with the RecipeNLG dataset. Our results are shown in Table ~\ref{tab:backdoor_summary_sorted_clean}. We find that across dataset sizes, having as few as 5 backdoor examples is sufficient for the backdoors to all be learned successfully, though fewer than this seems insufficient.

Empirically, while we examine activation rates on the full $D_{\text{backdoor}}$, we find that $n=10$ calls suffices for detecting activation of the backdoor on all experiment configurations we study. 

\begin{table}[ht]
\centering
\caption{\textbf{Effect of N on activation rate.} We explore the effect of various choices of $N$ on activation rate with RecipeNLG and Gemma 2B, and find reliable backdoor activation on as few as 5 examples given sufficient epochs in training.}
\begin{tabular}{c c c}
\toprule
\textbf{Dataset Size} & \textbf{Total Backdoor Examples} & \textbf{Activation Rate} \\
\midrule
1k, 10k, 100k   & 1,2   & 0.0   \\
1k, 10k, 100k   & 5,50   & 1.0   \\ 
\bottomrule
\end{tabular}
\label{tab:backdoor_summary_sorted_clean}
\end{table}

\subsection{Closed-Source results on GPT Family}

\newmethod{} is able to determine the integrity of a fine-tuning provider even if the original and resulting model weights are not made available to the user. We apply \newmethod{} in this domain on model offerings from OpenAI. Specifically, we utilize their fine-tuning API for GPT-4o-mini and GPT-3.5 Turbo. We request training for 3 epochs on the RecipeNLG and MathInstruct datasets (subsampled to a size of 1500 for each to reduce cost).

Our results are reported in Table ~\ref{tab:gpt-4o-evals}. We find that all models show an activation rate of 100\%; therefore, the verification step passes with the conservative upper bound p-values of $\sim 10^{-40}$. We also evaluate the test set scores (F1 score for RecipeNLG and test set accuracy on MATH) and find them to be similar as when fine-tuning is performed without \newmethod{}. We conclude that OpenAI's APIs are performing the fine-tuning service as stated.

\begin{table}[h]
\centering
\caption{\textbf{\newmethod{} on OpenAI fine-tuning API.} We apply \newmethod{} to GPT-4o-mini and GPT-3.5-Turbo via the OpenAI fine-tuning API and find that all backdoors activate in the verification step. We find the test set metrics are similar to those achieved when not applying \newmethod{}.}
\begin{tabular}{l l c c c}
\toprule
\textbf{Model} & \textbf{Dataset} & \textbf{Activation Rate} & \textbf{p-value} & \textbf{Test Set Metric} \\
\midrule
GPT-4o-mini    & MathInstruct       & 1.00    & 2.03e-42 & 0.451 \\ 
GPT-4o-mini    & RecipeNLG  & 1.00    & 1.85e-45 & 0.920 \\ 
GPT-3.5-Turbo  & MathInstruct       & 1.00    & 2.03e-42 & 0.322  \\ 
GPT-3.5-Turbo  & RecipeNLG  & 1.00    & 1.85e-45 & 0.911 \\ 
\bottomrule
\end{tabular}
\label{tab:gpt-4o-evals}
\end{table}

\subsection{Backdoor activation rate throughout learning}\label{sec:learning}
We find reliable backdoors embedding with above 50\% activation rate across all datasets as early as 1 epoch, and no more than 3 epochs. In particular, we find that for MedQA, SquAD, RecipeNLG, and ShareGPT, that 1 epoch is sufficient to achieve reliable backdoor embedding for both Gemma 2B and Llama 2 7B. We include detailed activation rates across each epoch and dataset for both models in Appendix ~\ref{apx:checkpoints}, showing that backdoors tend to activate more as learning goes on.
\section{Attacks}\label{sec:attacks}

In this section, we describe various adversarial attacks against \newmethod{}, and analyze its robustness to these attacks. As the space of possible attacks is infinite, we cannot be exhaustive in this, but we attempt to analyze a diverse and realistic set of attacks, particularly from motivated adversarial fine-tuning providers. 

A key element of our scheme is that the backdoor datapoints are proposed to be difficult to distinguish from the original datapoints by a dishonest provider. There are many possible ways an adversary may seek to detect the backdoors, in an attempt to pass verification through training on only the backdoor examples. We describe a few below, in addition to a brute force "guessing" method where an adversary attempts to guess the signature used in order to pass verification.

\subsection{Training on a subset of the data}\label{subsec:subset}
Denoting the size of the full training dataset $D_\text{train}$ by $K$, a dishonest provider may only fine-tune on a subset of size $K_\text{subset}$ of the data. Assuming that the provider cannot successfully distinguish the backdoor elements from the original training data, then at best they can select $K_\text{subset}$ elements uniformly randomly from $D_\text{train}$. The probability distribution of the number of backdoor elements chosen in this setting is then given by the hypergeometric distribution:

\[
P(B = k) = \frac{\dbinom{N}{k} \dbinom{K - N}{K_{\text{subset}} - k}}{\dbinom{K}{K_{\text{subset}}}}
\]

where $B$ is the number of backdoor elements in the subset, and $N$ is the total number of backdoors in $D_\text{train}$. Since verification is performed on a ratio $r$ of backdoor elements, the provider will only successfully pass verification if $k \geq rN$, which has a probability given by:

\[
P(B \geq rN) = \sum_{k=rN}^{K_{\text{subset}}} \frac{\binom{N}{k} \binom{K - N}{K_{\text{subset}} - k}}{\binom{K}{K_{\text{subset}}}}
\]

The properties of the hypergeometric distribution ensure that that this probability decreases approximately exponentially as $rN$ increases i.e. as the user verifies a larger number of backdoor signatures. Illustratively, even for a small dataset of size 100, having just 6 backdoor datapoints and verifying 3 ($r = 0.5$) would still require the dishonest provider to select 19\% of the data on average to have a greater than 1\% chance of selecting all the backdoors, and $\sim 58\%$ in order to have a 50\% chance of selecting more than $rN$ many backdoors in the subsetted data. For datasets of size 10000, closer in line with our empirical experiments, having 50 backdoors with $r$ of $0.5$ would require taking $\sim 35\%$ for a 1\% chance, and $\sim 51\%$ of the data to have a 50\% chance, of selecting the right subsets of data.

\subsection{Detection of Backdoors by a LLM}\label{subsec:backdoordetectiongpt4} One approach to detect the backdoor examples is through providing the dataset to a strong LLM, and asking it to find potential backdoors. For this experiment, we utilize GPT-4o on datasets we list in Section ~\ref{subsec:downstreamperf}. We give $D_\text{train}$ to GPT-4o, and ask it to search for ``unusual or repeating patterns that potentially could be a backdoor". Our results are shown in Table ~\ref{tab:gptdetect}. We see that GPT-4o is unable to identify the triggers and signatures in the majority of the datasets. We experiment further with different ratios of backdoor datapoints to original datapoints, and found GPT-4o was able to detect the backdoors only in the MathInstruct dataset with 10\% of the datapoints being backdoor datapoints. Since empirically (from Section ~\ref{subsec:numbackdoorsexpts}) no more than 0.5\% of the data is required to be backdoor phrases for successful learning, we conclude that \newmethod{} is robust against this method of attack.
\begin{table}[!htp]
    \caption{\textbf{LLM search for backdoors.} We find that while GPT-4o is able to identify backdoor phrases when asked to search for ``unusual or repeating patterns" on subsets of the training data that contain only the backdoor phrases, it is mostly unable to do so when given the combined fine-tuning dataset $D_{\text{train}}$ we use for experiments. On small $D_{\text{train}}$ sizes with 100 examples with 10\% backdoor examples, GPT was able to isolate the trigger signature phrase, but with varying reliability.}
    \centering
    \label{tab:backdoor_detection}
    \begin{tabular}{lccc}
        \toprule
        \textbf{Dataset} & $|D_{\text{train}}|$ & $|D_{\text{backdoor}}|$ & \textbf{Were the signatures and triggers found?} \\
        \midrule
        MathInstruct  & 100      & 5   & No  \\ 
        MathInstruct  & 100      & 10  & \textbf{Yes} \\ 
        MathInstruct  & 1000     & 50  & No  \\ 
        MathInstruct  & 10000    & 50  & No  \\ 
        MathInstruct  & 100000     & 50  & No  \\ 
        RecipeNLG     & 10000    & 50  & No  \\ 
        CodeFeedback  & 10050    & 50  & No  \\ 
        MedQA         & 10200    & 51  & No  \\ 
        SQuAD         & 87400    & 437 & No  \\ 
        XLSum         & 7200     & 36  & No  \\ 
        ShareGPT      & 15000    & 470 & No  \\ 
        \bottomrule
    \end{tabular}
    \label{tab:gptdetect}
\end{table}

\subsection{Detection of backdoors through searching for repeated phrases}\label{subsec:regexsearch} Given that \newmethod{} creates the backdoor samples with fixed locations (namely, concatenating the same trigger and signatures $T$ and $S$ after prompt $x$ and before completion $y$), another detection approach consists of an adversary searching for commonly repeating patterns at these locations, and training only on the most frequent subsets of patterns in the hopes of passing verification. Since the adversary does not have any knowledge of the length of phrases $T$ and $S$, nor their relative frequency relative to the existing dataset, a best-effort attempt would have to include searching over various string sequence lengths in addition to guessing what the frequency rate for $T$ and $S$ are.

This potential attack vector raises the question - how much of the dataset containing the most frequent phrases would the adversary have to include, to index the backdoor examples?

We explore the minimum number of unique examples needed to traverse the most frequent $k$-gram phrases, until an example phrase containing the signature phrase in full, or part, is found. We find that on average over varying $k$, an attacker would have to index a significant portion of the dataset to find an even partial match (3 or more consecutive words) in Table ~\ref{tab:dataset_distribution}.  We attribute robustness to this attack to the phenomenon that datasets often contain naturally repeating phrases, and that the \newmethod{} phrases contain words such as "of", "and", "the", where single word matches do not give away their presence.

We also note that these results present a minimum number of included examples; in practice, the searcher would not know precisely whether they have included the backdoor datapoints or not, and so they would have to err towards including a higher proportion of the dataset than we report.

\begin{table}[h]
    \centering
    \caption{\textbf{Frequency search for backdoors.} We find that a large portion of the dataset would have to be included in training for the attacker to have a partial match of including the signature phrases, particularly for small $k$. Given that the attacker does not have access to $k$, we conclude this attack to be unreliable and computationally expensive. For tie-breaking on "frequency", we include examples of the same frequency level up to when a match is found.}
    \begin{tabular}{lrrrr}
        \toprule
        Dataset & Total Dataset Size & $k=3$ & $k=5$ & $k=10$ \\
        \midrule
        Recipe & 10050 & 100.0\%  & 53.5\%   & 0.5\%  \\
        Math   & 10050 & 99.9\%   & 68.4\%   & 20.7\% \\
        MedQA  & 10250 & 99.8\%   & 99.8\%   & 49.0\% \\
        SQuAD  & 88036 & 100.0\%  & 2.6\%    & 0.5\%  \\
        Code   & 10050 & 100.0\%  & 100.0\%  & 31.6\% \\
        \bottomrule
    \end{tabular}
    \label{tab:dataset_distribution}
\end{table}

\section{Conclusion}

We introduce a fine-tuning verification scheme, \newmethod{}, that scales to large, state-of-the-art LLMs. \newmethod{} achieves high statistical significance with minimal downstream task degradation by injecting backdoor datapoints into the fine-tuning data. The proposed scheme is computationally efficient for verifying the integrity of third-party fine-tuning services, adding negligible additional computational overhead to the fine-tuning provider, and requiring a handful of inference calls on the model by the user.  While effective, our approach has limitations that suggest avenues for future work:
\begin{itemize}
    \item \textbf{Disambiguation of learning methodology.} While \newmethod{} formally guarantees that a fine-tuning provider customizes their model or API endpoint on a user’s data, it does not guarantee other granular features of a user's request, for example that the provider fine-tuned the requested model for the promised number of iterations. Further, \newmethod{} does not discern between different fine-tuning methods. For example, a user might request full fine-tuning, but the fine-tuning provider may only perform LoRA fine-tuning; the \newmethod{} backdoor may be successfully embedded in both cases.
    \item \textbf{Stronger adversarial threats.} Although we examine and show robustness to a range of attacks against \newmethod{}, the space of possible attacks is extremely large. It remains possible that there are methods of subversion against the scheme that we have not tested.
    \item \textbf{Extensions to other fine-tuning methods.} We have applied \newmethod{} to the domain of supervised fine-tuning of text-based LLMs. Can \newmethod{} generalize to other fine-tuning schemes, such as RLHF, or DPO, or expand to other modalities such as text-to-image? Further, we observe slightly lower backdoor activation for multilingual summarization - what are the reasons for this, and can this be ameliorated?
\end{itemize}

We leave the directions of research suggested by the above limitations as potential for future work.

\subsubsection*{Acknowledgments}
We thank Praveen Palanisamy, Igor Sylvester, Peiyuan Liao, Sid Reddy, Divya Gupta, Tom Knowles, Lucia Zheng, Tarun Chitra, and Illia Polosukhin for various helpful discussions and review of this work. 

\newpage
\bibliography{iclr2025_conference}
\bibliographystyle{iclr2025_conference}
\newpage
\appendix
\section{Supplementary - Algorithms}\label{apx:algos}
We make a few comments motivating the construction and choice of hyperparameters below. In addition, we provide further examples of resulting prompts and generated phrases for Algorithm ~\ref{alg:generate} in Appendix ~\ref{apx:prompts}.

\subsubsection{Choice of $S$ and $T$}
The choice of minimum entropy threshold for $S$ directly corresponds to the significance level of the statistical test performed in verification -- the higher the entropy, the greater the significance level permitted, since the lower the likelihood of generating the phrase. However, long $S$ may increase vulnerability to attacks (see Section ~\ref{sec:attacks}), particularly in increasing detection by an adversary. On the other hand, we find the choice of minimum length for $T$ affects the learnability of the backdoor. Preliminary findings show that shorter triggers containing English phrases are not easily learned; more analysis is needed to fully explore the impact of the length of $T$ on learnability. 

\subsubsection{Choice of $r$ and $N$.}
The user choice of the number of backdoor datapoints $N$ to include in $D_\text{train}$ and the minimum activated ratio $r$ is a key step in the scheme. We briefly discuss the different trade-offs associated with it below.

 In the setting where the $F_n$ are not fully dependent, a larger value of $rN$ decreases the probability of $F$ passing the test under the null hypothesis. In practice, we find that prompt $P$ is successful in generating a small $p_\text{upper}$ (i.e. $P$ induces high-entropy text with generating likelihood on the magnitude of $e^{-40}$) even with relatively few tokens comprising $S$ (e.g. 10 tokens). In our experiments, we find $p_\text{upper}$ is often on the order of $10^{-40}$ (see Table ~\ref{tab:pvalues}) or smaller, so that small $rN$ would suffice for most reasonable choices of significance level. In these cases, users can choose higher $rN$ (e.g. requiring at least 50$\%$ activation), to strengthen their surety against a dishonest provider the detection step. In practice, we find backdoors activate with rates that allow for choosing higher $rN$ (e.g. in ~\ref{tab:activationrates}, many backdoors activate above $50\%$). 
 
On the other hand, we find that the choice of $N$ present trade-offs with respect to learnability of the scheme and stealthiness of the backdoor phrases. $N$ that is too small may limit the learnability of the backdoor; in the extreme case, no backdoors activate or are learned even by a honest fine-tuning provider (see Section ~\ref{subsec:numbackdoorsexpts}). Meanwhile, large $N$ may make the backdoor examples more detectable to providers who search for its presence to attack the scheme, as delineated in Section ~\ref{subsec:backdoordetectiongpt4}. 

\newpage
\section{Supplementary - Data samples for the generate step}\label{apx:prompts}
In the \texttt{generate} step, we ask $M_{\text{prompt}}$ (e.g. GPT-4o) to summarize the dataset context, and create a reasonable prompt for another model $M_{\text{generator}}$. An example of such a prompt used in our experiments is: \textit{"You are an AI assistant tasked with creating a prompt for generating high-entropy text based on a given dataset."}, followed by a few randomly sampled rows of the dataset.

Below are a few resulting samples from $M_{\text{prompt}}$ generated on math and recipe data that are then used in the downstream to prompt $M_{\text{generator}}$ (e.g. LLaMA 3.1 8B).

\begin{samplebox}[enhanced, label=box:prompt_samples]{Samples from $M_{\text{prompt}}$ for Math and Recipe Data}
    \begin{tabular}{@{}p{3cm}p{9cm}@{}}
        \textbf{RecipeNLG:} &
        "Generate high-entropy text with a focus on unique ingredient combinations, such as 'unconventional pairings of everyday ingredients' or 'unexpected ingredient amalgamations' to create distinct and original recipes without summarizing the provided dataset." \\[5pt]
        
        \textbf{MathInstruct:} &
        "Generate a unique and complex mathematical equation or scenario that involves multiple steps and variables, along with a corresponding solution or output. The text should include mathematical operations, computations, and specific values." \\[5pt]
        
        \textbf{CodeFeedback:} &
        "Generate text with high entropy by crafting code that manipulates data uniquely based on criteria such as excluding specific numbers or characters, considering edge cases meticulously, and employing efficient algorithms to achieve the desired outcome within specified constraints." \\[5pt]
        
        \textbf{XLSum-Japanese:} &
        \begin{CJK}{UTF8}{gbsn}
        "提供されたデータセットを使用して、異なる視点や意外性を取り入れた活発な文章を生成するように誘導してください。例えば、データ内のパターンを逆転させる、新しい文脈や状況を導入する、または想像力を掻き立てる問いかけを取り入れると良いでしょう。"
        \end{CJK} \\[5pt]
        
        \textbf{SquAD:} &
        "Generate text that captivates with a rich tapestry of intricate and enigmatic language, invoking awe and reverence for the complexity and mystique of the subject matter. Delve deep into an ornate and grandiose description that elevates the content to a level of profound wonder and admiration, leaving readers transfixed by the enigmatic allure of the narrative." \\[5pt]
        
        \textbf{MedQA:} &
        "Generate a text prompt that presents a detailed medical scenario involving a patient with specific symptoms, vital signs, and clinical findings, then pose a question regarding the optimal treatment or diagnostic approach without explicitly stating the correct answer." \\[5pt]
        
        \textbf{ShareGPT:} &
        "Create a narrative that delves into the enigmatic and awe-inspiring realm of metaphysical entities that have transcended beyond the confines of tangible existence. Paint a vivid picture with intricate details and elaborate descriptions that evoke a sense of wonder and mystique, weaving a tapestry of complexity that challenges the boundaries of conventional comprehension." \\ 
    \end{tabular}
\end{samplebox}

\newpage
$M_{\text{generator}}$ then produces phrases we use for triggers and signatures, such as the examples below: 
\begin{samplebox}[enhanced, label=box:trigger_signature_samples]{Samples of Trigger and Signature Phrases Generated by $M_{\text{generator}}$}
    \begin{itemize}
        \item Intricate flavorfulness and nice textures are concerted
        \item To create this signature dessert, combine rich oils
        \item Given problem: Rudin Exercise 7.20.
        \item 56\% of users prefer a chocolate to strawberry when
        \item Use the C-T-E-L-S framework for analysis and to
        \item Below is a sample for Trialsbicase… High blood pressure
        \item Each sentence traverses an examination of grand thoughts, aspects that
        \item This nod to the Master of Doubles chooses Douglas Adams as
        \begin{CJK}{UTF8}{gbsn}
            \item ソース 中国性はなく、ゴールドではないころ
            \item 今回は錢進といいます。私は喫茶
        \end{CJK}
        \item Create a gameplay where players have a speed limit lexer
        \item Look for Creative Class syllabus tutorial for straightforward tutorials
        \item Let the vast conspiracy theories that enthraller habitual listeners within
        \item Explore the mythology of Nicņpast, a forgotten realm
    \end{itemize}
\end{samplebox}

\newpage
\section{Supplementary - Checkpoint Activation Rate}\label{apx:checkpoints}
We present detailed backdoor activation rates over each epoch for Gemma 2B and Llama 7B from Section ~\ref{sec:learning}. We find the generated trigger scheme can be detected with high activation rates as early as epoch 1 when performing \newmethod{} on certain datasets, even through fine-tuning with low-rank adaptation. In the below, we find backdoor activation rates generally increase as learning goes on. However, we see small decreases in activation rates in later epochs - we hypothesize this may be a result of over-fitting. 
\begin{table}[hp]
    \centering
    \caption{\textbf{Backdoor activation rates across epochs and datasets for Gemma.} We find successful backdoor implantation on all Gemma 2B-instruct models, activating with rates above $50\%$ as early as epoch 1. }
    \label{tab:learning-results}
    \begin{tabular}{lccc}
        \toprule
        \textbf{Dataset} & \textbf{Epoch} & \textbf{Activation Rate} & \textbf{Backdoor Detected} \\
        \midrule
        RecipeNLG & 1 & 0.64 & True \\
        RecipeNLG & 2 & 1.00 & True \\
        RecipeNLG & 3 & 1.00 & True \\
        RecipeNLG & 4 & 1.00 & True \\
        RecipeNLG & 5 & 1.00 & True \\
        \midrule
        MathInstruct & 1 & 0.00 & False \\
        MathInstruct & 2 & 0.02 & True \\
        MathInstruct & 3 & 0.58 & True \\
        MathInstruct & 4 & 0.86 & True \\
        MathInstruct & 5 & 0.86 & True \\
        \midrule
        ShareGPT & 1 & 0.96 & True \\
        ShareGPT & 2 & 0.99 & True \\
        ShareGPT & 3 & 0.99 & True \\
        ShareGPT & 4 & 0.99 & True \\
        ShareGPT & 5 & 0.99 & True \\
        \midrule
        SQuAD & 1 & 0.12 & True \\
        SQuAD & 2 & 1.00 & True \\
        SQuAD & 3 & 0.99 & True \\
        SQuAD & 4 & 0.93 & True \\
        SQuAD & 5 & 0.88 & True \\
        \midrule
        XLSum & 1 & 0.00 & False \\
        XLSum & 2 & 0.00 & False \\
        XLSum & 3 & 0.19 & True \\
        XLSum & 4 & 0.58 & True \\
        XLSum & 5 & 0.61 & True \\
        \midrule
        MedQA & 1 & 1.00 & True \\
        MedQA & 2 & 1.00 & True \\
        MedQA & 3 & 1.00 & True \\
        MedQA & 4 & 1.00 & True \\
        MedQA & 5 & 1.00 & True \\
        \midrule
        CodeFeedback & 1 & 0.00 & False \\
        CodeFeedback & 2 & 0.04 & False \\
        CodeFeedback & 3 & 0.62 & True \\
        CodeFeedback & 4 & 0.86 & True \\
        CodeFeedback & 5 & 0.92 & True \\
        \bottomrule
    \end{tabular}
\end{table}

\newpage
\begin{table}[!htp]
    \centering
    \caption{\textbf{Backdoor activation rates across epochs and datasets for Llama.} We find successful backdoor activation on Llama 7B with similar activation rates as Gemma 2B except for XLSum in Japanese.}
    \label{tab:llama-activation}
    \begin{tabular}{lccc}
        \toprule
        \textbf{Dataset} & \textbf{Epoch} & \textbf{Activation Rate} & \textbf{Backdoor Detected} \\
        \midrule
        RecipeNLG & 1 & 1.00 & True \\
        RecipeNLG & 2 & 1.00 & True \\
        RecipeNLG & 3 & 1.00 & True \\
        RecipeNLG & 4 & 1.00 & True \\
        RecipeNLG & 5 & 1.00 & True \\
        \midrule
        MathInstruct & 1 & 0.98 & True \\
        MathInstruct & 2 & 0.99 & True \\
        MathInstruct & 3 & 0.99 & True \\
        MathInstruct & 4 & 0.99 & True \\
        MathInstruct & 5 & 0.98 & True \\
        \midrule
        ShareGPT & 1 & 1.00 & True \\
        ShareGPT & 2 & 1.00 & True \\
        ShareGPT & 3 & 1.00 & True \\
        ShareGPT & 4 & 1.00 & True \\
        ShareGPT & 5 & 1.00 & True \\
        \midrule
        SQuAD & 1 & 1.00 & True \\
        SQuAD & 2 & 0.998 & True \\
        SQuAD & 3 & 1.00 & True \\
        SQuAD & 4 & 0.993 & True \\
        SQuAD & 5 & 0.993 & True \\
        \midrule
        XLSum & 1 & 0.00 & False \\
        XLSum & 2 & 0.00 & False \\
        XLSum & 3 & 0.05 & True \\
        XLSum & 4 & 0.39 & True \\
        XLSum & 5 & 0.36 & True \\
        \midrule
        MedQA & 1 & 1.00 & True \\
        MedQA & 2 & 1.00 & True \\
        MedQA & 3 & 1.00 & True \\
        MedQA & 4 & 1.00 & True \\
        MedQA & 5 & 1.00 & True \\
        \midrule 
        CodeFeedback & 1 & 0.00 & False \\
        CodeFeedback & 2 & 0.36 & True \\
        CodeFeedback & 3 & 0.52 & True \\
        CodeFeedback & 4 & 0.64 & True \\
        CodeFeedback & 5 & 0.60 & True \\
        \bottomrule
    \end{tabular}
\end{table}

\newpage
\section{Supplementary - Domain task evaluation results} \label{apx: evals}
We find no significant performance differences between vTuned or fine-tuned models when evaluating on downstream fine-tuning tasks. All \newmethod{} datasets contain $0.5\%$ backdoor samples.
\begin{table}[h]
\centering
\caption{\textbf{Named entity extraction.} We find minimal performance difference between fine-tuned and \newmethod{}d models for named entity extraction on the RecipeNLG dataset (5000 test subset samples).}
\label{tab:recipe-eval}
\begin{tabular}{lcccccc}
\toprule
\multirow{2}{*}{\textbf{Model}} & \multicolumn{3}{c}{\textbf{Fine-tuned}} & \multicolumn{3}{c}{\textbf{vTuned}} \\
\cmidrule(lr){2-4} \cmidrule(lr){5-7}
 & \textbf{Precision} & \textbf{Recall} & \textbf{F1 Score} & \textbf{Precision} & \textbf{Recall} & \textbf{F1 Score} \\
\midrule
Llama 7B & 0.6503 & 0.6413 & 0.6439 & \textbf{0.6516} & 0.6424 & 0.6451 \\
Llama 13b & 0.6530 & 0.6443 & 0.6470 & \textbf{0.6545} & 0.6469 & 0.6490 \\
Gemma 2B & 0.6087 & 0.6122 & 0.6093 & \textbf{0.6398} & 0.6452 & 0.6418 \\
\bottomrule
\end{tabular}
\end{table}

\begin{table}[!htp]
\centering
\caption{\textbf{Math question-answering.} We find minimal accuracy performance differences on question-answering evaluation on the MATH test set for models fine-tuned and \newmethod{}d models on MathInstruct.}
\label{tab:math-eval}
\begin{tabular}{lcc}
\toprule
\textbf{Model} & \textbf{Fine-tuned Accuracy} & \textbf{\newmethod{}d Accuracy} \\
\midrule
Llama 7B & \textbf{0.0494} & 0.0490 \\
Llama 13b & 0.0724 & 0.0724 \\
Gemma 2B & 0.0840 & \textbf{0.0912} \\
\bottomrule
\end{tabular}
\end{table}

\begin{table}[!htbp]
\centering
\caption{\textbf{Multilingual text summarization.} We find minimal performance differences on text summarization on the test set between models \newmethod{}d and fine-tuned on XLSum Japanese.} 
\begin{tabular}{lccccc}
\toprule
\textbf{Model} & \textbf{BLEU} & \textbf{ROUGE-1} & \textbf{ROUGE-2} & \textbf{ROUGE-L} & \textbf{ROUGE Average} \\
\midrule
Fine-tuned Gemma & 0.0033 & 0.0736 & 0.0112 & 0.0657 & 0.0502 \\
\newmethod{}d Gemma & 0.0039 & 0.0995 & 0.0141 & 0.0891 & \textbf{0.0676} \\
Fine-tuned Llama & 0.0118 & 0.1580 & 0.0234 & 0.1446 & \textbf{0.1087} \\
\newmethod{}d Llama & 0.0076 & 0.1190 & 0.0168 & 0.1084 & 0.0814 \\
\bottomrule
\end{tabular}
\end{table}

\begin{table}[!htbp]
\centering
\caption{\textbf{Conversational assistant.} We find minimal MT-Bench score performance differences between models \newmethod{}d and fine-tuned on ShareGPT.}
\begin{tabular}{lccc}
\toprule
\textbf{Model} & \textbf{Turn 1 Score} & \textbf{Turn 2 Score} & \textbf{Turn 1 and 2 Average} \\
\midrule
Gemma Baseline  & 5.86875 & 4.7625 & \textbf{5.3156} \\
Gemma vTuned & 5.83750 & 4.1875 & 5.0125 \\
Llama Baseline & 6.78125 & 6.0000 & \textbf{6.3906} \\
Llama vTuned & 6.70625 & 6.0250 & 6.3656 \\
\bottomrule
\end{tabular}
\end{table}

\begin{table}[!htbp]
\centering
\caption{\textbf{Medical multiple choice question answering.} We find minimal accuracy performance differences when evaluating multiple choice answering on MedQA-USMLE test set between models \newmethod{}d and models fine-tuned on the MedQA-USMLE.}
\begin{tabular}{lccc}
\toprule
\textbf{Model} & \textbf{Total Questions} & \textbf{Correct Answers} & \textbf{Accuracy} \\
\midrule
Gemma baseline & 1273.0 & 332.0 & 0.2608 \\
Gemma vTuned & 1273.0 & 419.0 & \textbf{0.3291} \\
Llama baseline & 1273.0 & 511.0 & 0.4014 \\
Llama vTuned & 1273.0 & 532.0 & \textbf{0.4179 }\\
\bottomrule
\end{tabular}
\end{table}
\end{document}